\documentclass[runningheads]{llncs}
\usepackage{graphicx}
\usepackage{tikz}
\usepackage{comment}
\usepackage{amsmath,amssymb} % define this before the line numbering.
\usepackage{color}
\usepackage{graphicx}
\usepackage{caption}
\usepackage{subcaption}
\usepackage[pagebackref,breaklinks,colorlinks]{hyperref}
\usepackage{booktabs}

\newcommand{\ie}{\textit{i}.\textit{e}. }
\newcommand{\eg}{\textit{e}.\textit{g}. }
\definecolor{demphcolor}{RGB}{144,144,144}

\newcommand\blfootnote[1]{%
\begingroup
\renewcommand\thefootnote{}\footnote{#1}%
\addtocounter{footnote}{-1}%
\endgroup
}

\usepackage[accsupp]{axessibility}  % Improves PDF readability for those with disabilities.

\begin{document}
% \renewcommand\thelinenumber{\color[rgb]{0.2,0.5,0.8}\normalfont\sffamily\scriptsize\arabic{linenumber}\color[rgb]{0,0,0}}
% \renewcommand\makeLineNumber {\hss\thelinenumber\ \hspace{6mm} \rlap{\hskip\textwidth\ \hspace{6.5mm}\thelinenumber}}
% \linenumbers
\pagestyle{headings}
\mainmatter
\def\ECCVSubNumber{6672}  % Insert your submission number here

\title{Pose for Everything: Towards Category-Agnostic Pose Estimation} % Replace with your title

% INITIAL SUBMISSION 
\begin{comment}
\titlerunning{ECCV-22 submission ID \ECCVSubNumber} 
\authorrunning{ECCV-22 submission ID \ECCVSubNumber} 
\author{Anonymous ECCV submission}
\institute{Paper ID \ECCVSubNumber}
\end{comment}
%******************

% CAMERA READY SUBMISSION
% \begin{comment}
\titlerunning{Pose for Everything}
% If the paper title is too long for the running head, you can set
% an abbreviated paper title here
%
\author{Lumin Xu\inst{1,2*} \and Sheng Jin\inst{3,4*} \and Wang Zeng\inst{1,2} \and	Wentao Liu\inst{4,5} \and	Chen Qian\inst{4} \\ Wanli Ouyang\inst{5,6} \and Ping Luo\inst{3} \and Xiaogang Wang\inst{1} \\[.21cm]
	$^{1}$ The Chinese University of Hong Kong \quad
	$^{2}$ SenseTime Research \\
	$^{3}$ The University of Hong Kong \quad
	$^{4}$ SenseTime Research and Tetras.AI  \\
	$^{5}$ Shanghai AI Laboratory \quad
	$^{6}$ The University of Sydney \\
	\tt\small \{luminxu, zengwang\}@link.cuhk.edu.hk \quad js20@connect.hku.hk \\
	\{liuwentao, qianchen\}@sensetime.com \quad wanli.ouyang@sydney.edu.au \quad pluo@cs.hku.hk \quad xgwang@ee.cuhk.edu.hk}
\authorrunning{L. Xu et al.}
% First names are abbreviated in the running head.
% If there are more than two authors, 'et al.' is used.
%
\institute{}
% \end{comment}
%******************
\maketitle

\begin{abstract}

Existing works on 2D pose estimation mainly focus on a certain category, \eg human, animal, and vehicle. However, there are lots of application scenarios that require detecting the poses/keypoints of the unseen class of objects. 
In this paper, we introduce the task of Category-Agnostic Pose Estimation (CAPE), which aims to create a pose estimation model capable of detecting the pose of any class of object given only a few samples with keypoint definition.
To achieve this goal, we formulate the pose estimation problem as a keypoint matching problem and design a novel CAPE framework, termed POse Matching Network (POMNet). A transformer-based Keypoint Interaction Module (KIM) is proposed to capture both the interactions among different keypoints and the relationship between the support and query images.
We also introduce Multi-category Pose (MP-100) dataset, which is a 2D pose dataset of 100 object categories containing over 20K instances and is well-designed for developing CAPE algorithms. 
Experiments show that our method outperforms other baseline approaches by a large margin.
Codes and data are available at \url{https://github.com/luminxu/Pose-for-Everything}.
 
\keywords{2D pose estimation, class-agnostic, few-shot, MP-100 dataset}

\blfootnote{* indicates equal contribution.}

\end{abstract}

\section{Introduction}

2D pose estimation (also referred to as keypoint localization) aims to predict the locations of the pre-defined semantic parts of an instance. It has received great attention in the computer vision community in recent years because of its broad application scenarios in both academia and industry. For example, human pose estimation~\cite{mpii} has been widely used in virtual reality (VR) and augmented reality (AR); animal pose estimation~\cite{yu2021ap} is of great significance in zoology and wildlife conservation; vehicle pose estimation~\cite{reddy2018carfusion} is critical for autonomous driving. 

The real-world applications from different fields often involve detecting the poses of a variety of novel objects of interest. 
For example, biologists may study the plant growth by analyzing the poses of plants. However, traditional pose estimators are category-specific and can only be applied to the category that they are trained on. In order to detect poses of novel objects, users have to collect a huge amount of labeled data and design category-specific pose estimation models, which is time-consuming and laborious. To make matters worse, data collection for rare objects (\eg endangered animals) and semantic keypoint annotation for cases that need domain knowledge (\eg medical images) are extremely challenging. Therefore, there is increasing demand for developing pose estimation approaches that can generalize across different categories.

\begin{figure}[t]
	\centering
	\includegraphics[width=0.9\textwidth]{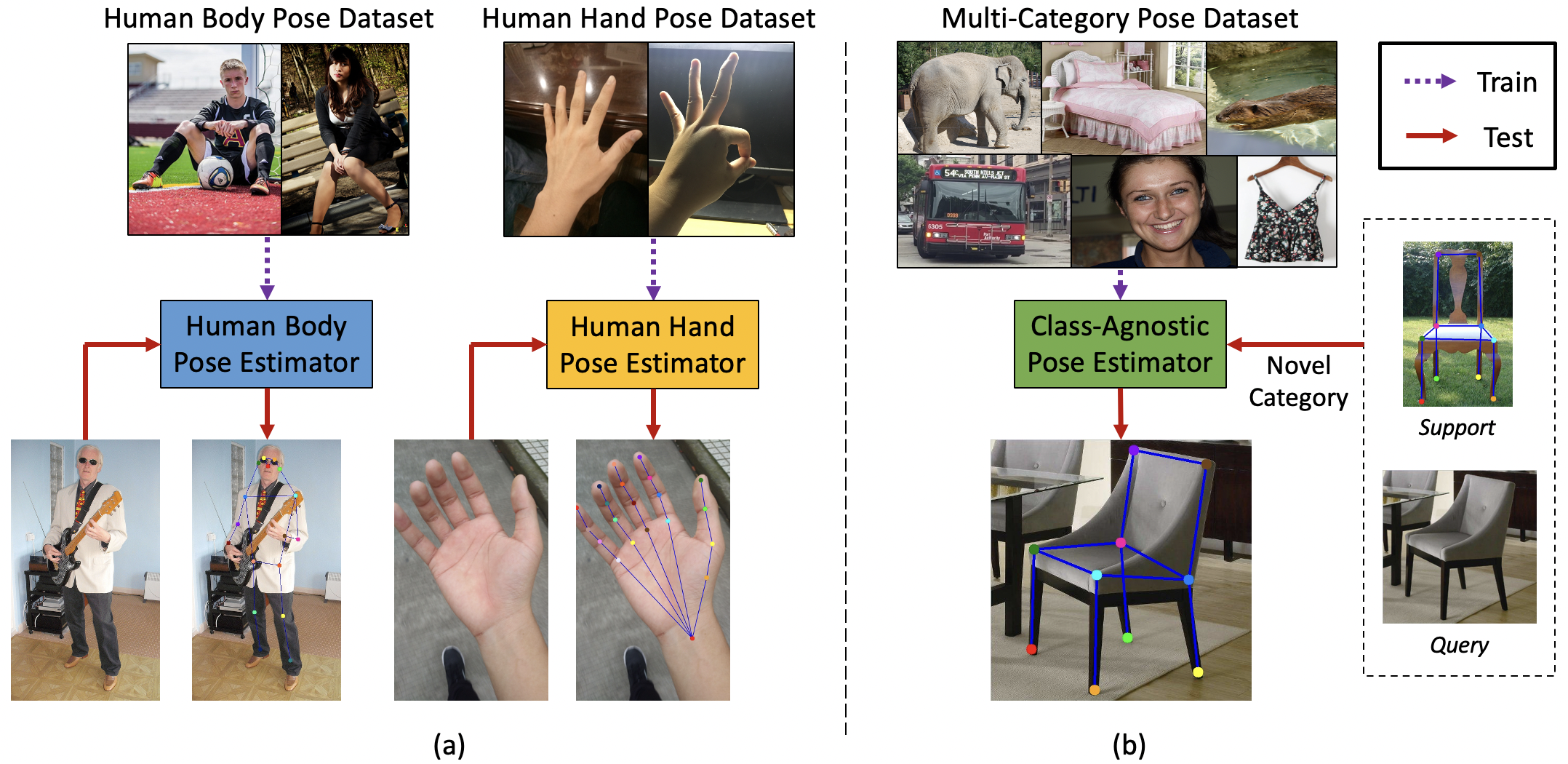}
	\caption{Category-Specific Pose Estimation vs Class-Agnostic Pose Estimation (CAPE). (a) Traditional pose estimation task is category-specific. Pose estimators are trained on the dataset containing objects of a single category, and can only predict the poses of that category. (b) CAPE task requires the pose estimator to detect poses of arbitrary category given the keypoint definition. After training on the pose dataset containing multi-category objects, the pose estimators can generalize to novel categories given one or a few support images.}
	\label{fig:intro}
\end{figure}

In this paper, we introduce an important yet challenging task, termed Category-Agnostic Pose Estimation (CAPE). As shown in Fig.~\ref{fig:intro}, unlike traditional pose estimation methods that can only predict the poses of a specific category, CAPE aims at using a single model for detecting poses of any category. Given a support image of a novel category and the corresponding keypoint definition, the class-agnostic pose estimator predicts the pose of the same category in the query image. In this way, the pose of any object of interest can be generated according to the arbitrary keypoint definition. The huge cost of data collection, model training and parameter tuning for each novel class is greatly reduced.

There are several challenges preventing the computer vision community from designing systems capable of predicting the poses of a large number of object categories. 
First, most pose estimation approaches~\cite{sun2019deep} treat it as a supervised regression task, requiring thousands of labeled images to learn to map an input image to keypoint locations. 
Second, different objects may have different keypoint definition and unknown number of keypoints. It is non-trivial to learn the unique output representations and utilize the structural information.
Third, there are few to none large-scale pose estimation datasets with many visual categories for the development of a general pose estimation method. Previous datasets mostly consist of only one category (\eg human body).

In this paper, we take the first step towards CAPE and propose a novel framework, termed POse Matching Network (POMNet). POMNet formulates the 2D pose estimation task as a matching problem. The keypoint features are extracted from the support images based on the reference keypoint definition, and the image features are extracted from the query image. Matching Head (MH) is designed, which integrates the support keypoint features and the query image features, to estimate the keypoint positions with the maximal possibility. In this way, the model is agnostic to the object category and can be used for any number of keypoints. A transformer-based Keypoint Interaction Module (KIM) is also proposed to capture both the connections among different keypoints and the relationship between the support and query images. The features of different keypoints mutually interact with each other to learn their inherent structure for the given object category. The keypoint features are further aligned with the query image features for better matching. Experimental results show that our model significantly outperforms the other baseline models by a large margin.

In order to train and evaluate the class-agnostic pose estimators, we collect a large-scale pose dataset called Multi-category Pose (MP-100) dataset. The dataset contains over 20K instances, covering 100 sub-categories (\eg vinegar fly body, sofa, suv, and skirt) and 8 super-categories (\eg animal face, furniture, vehicle, and clothes). To our best knowledge, it is the first benchmark that contains the pose annotation of multiple visual (super-)categories.

The main contributions of our work are three-folds. 

\begin{itemize}

\item We introduce an important yet challenging task termed Category-Agnostic Pose Estimation (CAPE). CAPE requires the model to predict the poses of any objects given a few support images with keypoint definition.

\item We propose the novel CAPE framework, namely POse Matching Network (POMNet), and formulate the keypoint detection task as a matching problem. Keypoint Interaction Module (KIM) is proposed to capture both the keypoint-level relationship and the support-query relationship.

\item We build the first large-scale multi-(super-)category dataset for the task of CAPE, termed Multi-category Pose (MP-100), to boost the related research. 

\end{itemize}

\section{Related Works}

\subsection{2D Pose Estimation}

There are two types of keypoints in computer vision community. Semantic points are points with clear semantic meanings (\eg the left eye), while interest points are low-level points (\eg corner points).
2D pose estimation focuses on predicting the semantic points of objects, \eg human body parts~\cite{duan2019trb,jin2020whole,lin2014microsoft}, facial landmarks~\cite{bulat2017far}, hand keypoints~\cite{zimmermann2017learning}, and animal poses~\cite{cao2019cross}.
However, current pose estimation methods and datasets~\cite{ge2019deepfashion2,lin2014microsoft,yu2021ap} only focus on keypoints of a single super-category and can not support cross-category/unseen pose estimation.

\textbf{2D Pose Estimation Method.}
Existing methods can be classified into two categories: regression-based methods~\cite{li2021human,nie2019single,sun2017compositional,toshev2014deeppose} and heatmap-based methods~\cite{chen2018cascaded,cheng2020higherhrnet,chu2017multi,jin2019multi,jin2020differentiable,jin2017towards,li2019crowdpose,newell2016stacked,sun2019deep,wei2016convolutional,xiao2018simple}. Regression-based approaches directly map the image to keypoint coordinates. Such methods are flexible and efficient for real-time applications. However, they are vulnerable to occlusion and motion blur, resulting in inferior performance.
Heatmap-based approaches use likelihood heatmaps to encode the keypoint location. Because of excellent localization precision, heatmap-based methods are dominant in the field of 2D pose estimation. 
Recent works on pose estimation mostly focus on designing powerful convolutional neural networks~\cite{chen2018cascaded,chu2017multi,newell2016stacked,sun2019deep,wei2016convolutional,xiao2018simple,xu2021vipnas} or transformer-based architectures~\cite{li2021tokenpose,mao2021tfpose,yang2020transpose,yuan2021hrformer,zeng2022not}. However, they only focus on detecting the keypoints of object categories that appear during training. In comparison, our model is capable of detecting the keypoints of arbitrary objects of unseen classes. 

\begin{figure}[t]
\begin{minipage}{.6\linewidth}
	\centering
	\includegraphics[width=0.95\textwidth]{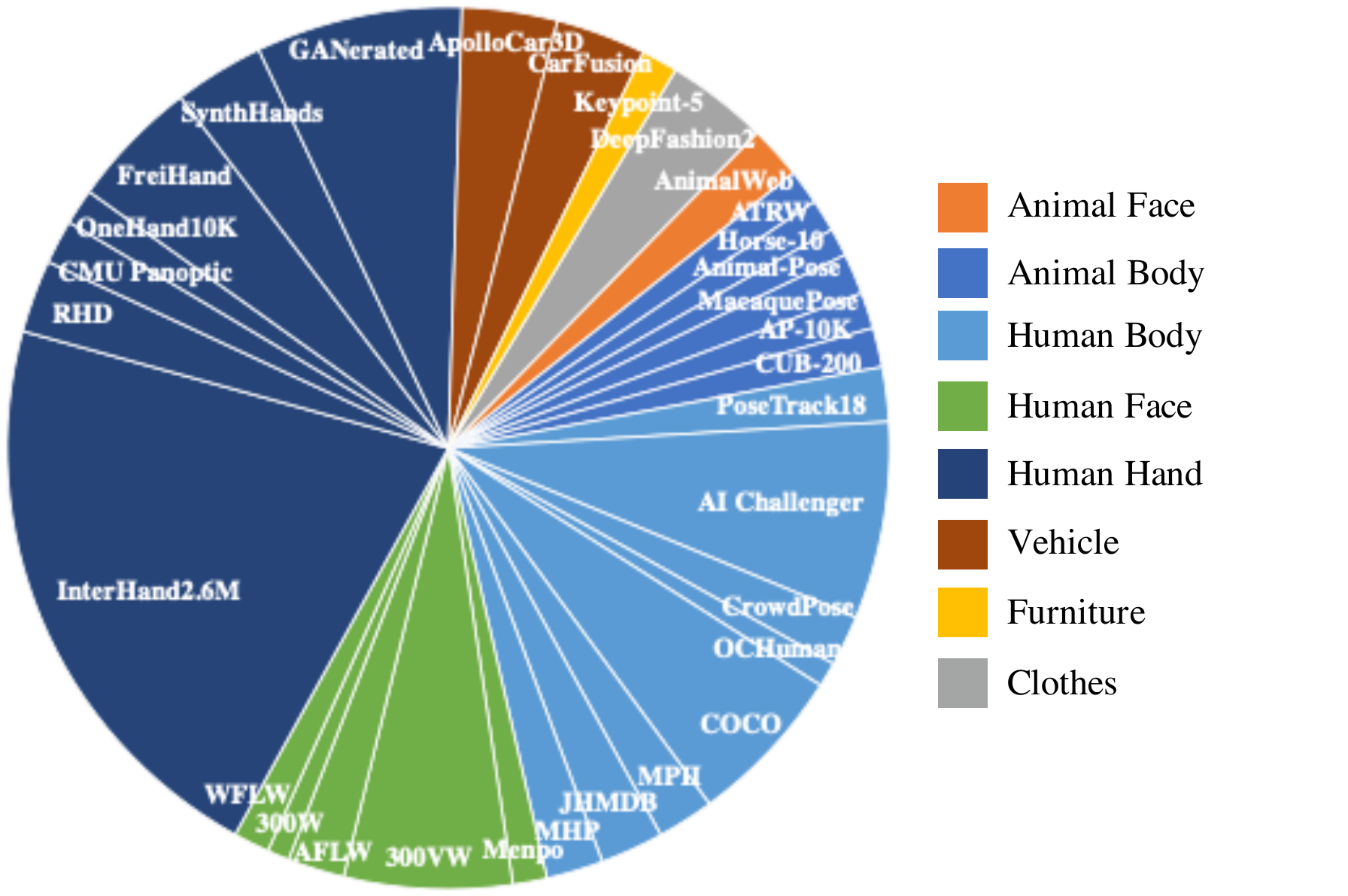}
\end{minipage}
\hspace{-8mm}
\begin{minipage}{.4\linewidth}
    \setlength\tabcolsep{4pt}
    \scalebox{0.6}{
		\begin{tabular}{cc|cc}
			\hline
			Dataset & \#Images & Dataset & \#Images \\ \hline
			AnimalWeb~\cite{khan2020animalweb} & 21.9K & 300VW~\cite{shen2015first} & 218K \\
			ATRW~\cite{li2020atrw} & 9.5K & AFLW~\cite{aflw} & 25K \\
			Horse-10~\cite{mathis2021pretraining} & 8.1K & 300W~\cite{300w} & 3.8K \\
			Animal-Pose~\cite{cao2019cross} & 6.1K & WFLW~\cite{wu2018look} & 10K \\
			MacaquePose~\cite{labuguen2021macaquepose} & 13K & InterHand2.6M~\cite{Moon_2020_ECCV_InterHand2.6M} & 2.6M \\
			AP-10K~\cite{yu2021ap} & 10K & RHD~\cite{zimmermann2017learning} & 41K \\
			CUB-200~\cite{cub-200-2011} & 12K & CMU Panoptic~\cite{simon2017hand} & 15K \\
			PoseTrack18~\cite{andriluka2018posetrack} & 23K & OneHand10K~\cite{wang2018mask} & 10K \\
			AI Challenger~\cite{wu2017ai} & 300K & FreiHand~\cite{zimmermann2019freihand} & 130K \\
			CrowdPose~\cite{li2019crowdpose} & 20K & SynthHands~\cite{mueller2017real} & 63K \\
			OCHuman~\cite{zhang2019pose2seg} & 4.7K & GANerated~\cite{mueller2018ganerated} & 330K \\
			COCO~\cite{lin2014microsoft} & 200K & ApolloCar3D~\cite{song2019apollocar3d} & 70K \\
			MPII~\cite{mpii} & 25K & CarFusion~\cite{reddy2018carfusion} & 63K \\
			JHMDB~\cite{Jhuang2013} & 31K & Keypoint-5~\cite{wu2016single} & 10K \\
			MHP~\cite{zhao2018understanding} & 25K & DeepFashion2~\cite{ge2019deepfashion2} & 80K \\
			Menpo~\cite{zafeiriou2017menpo} & 9K & & \\
			\hline
		\end{tabular}
	}
\end{minipage}
\caption{Categories and image numbers for popular 2D pose estimation datasets.}
\label{fig:all_dataset}
\end{figure}

\textbf{2D Pose Estimation Benchmark.}
Existing 2D pose estimation datasets only focus on a single super-category. As shown in Fig.~\ref{fig:all_dataset}, most attentions have been focused on human-related categories (\eg human body~\cite{andriluka2018posetrack,mpii,Jhuang2013,li2019crowdpose,lin2014microsoft,wu2017ai,zhang2019pose2seg,zhao2018understanding}, human face~\cite{aflw,300w,shen2015first,wu2018look,zafeiriou2017menpo}, and human hand~\cite{Moon_2020_ECCV_InterHand2.6M,mueller2018ganerated,mueller2017real,simon2017hand,wang2018mask,zimmermann2017learning,zimmermann2019freihand}), and there are numerous large-scale datasets for these classes. For other long-tailed categories, the datasets are relatively limited in terms of both the dataset sizes and diversity. Nevertheless, analyzing these long-tailed object categories is of great significance in both academia and industry. 
For example, vehicle pose estimation~\cite{reddy2018carfusion,song2019apollocar3d} is important for autonomous driving. 
Animal pose estimation~\cite{cao2019cross,labuguen2021macaquepose,li2020atrw,mathis2021pretraining,cub-200-2011,yu2021ap} is of great significance in zoology and wildlife conservation. 
Indoor furniture pose estimation~\cite{wu2016single} is important for developing household robots.
In this paper, we build the first large-scale benchmark (MP-100 dataset) that contains the pose annotations of a wide range of visual super-categories.

\subsection{Category-agnostic Estimation}
Category-agnostic estimation has been applied to many computer vision tasks, including detection~\cite{jiang2019class}, segmentation~\cite{zhang2019canet}, object counting~\cite{lu2018class,yang2021class} and viewpoint estimation~\cite{zhou2018starmap}. 
Our work is mostly related to StarMap~\cite{zhou2018starmap}, which proposes category-agnostic 3D keypoint representations encoded with canonical view locations. However, StarMap is only applicable for rigid objects (\eg furniture), and relies on several expensive 3D CAD models of the target category to identify the predicted keypoint proposals. In comparison, CAPE aims at predicting 2D poses of any object category (both rigid and flexible) according to any manual keypoint definition given by one or a few support images.

\subsection{Few-shot Learning}
Few-shot learning~\cite{lu2020learning} aims at learning novel classes using only a few examples. 
Recent few-shot learning approaches can be roughly classified into three categories, \ie metric-learning-based approaches~\cite{snell2017prototypical,vinyals2016matching,yang2018learning}, meta-learning-based approaches~\cite{finn2017model,ravi2016optimization}, and data-augmentation-based approaches~\cite{hariharan2017low}. 
\emph{Metric-learning-based Approaches.}
Prototypical networks~\cite{snell2017prototypical} learn the prototype (embedding features) of each class in the support data and then classify query data as the class whose prototype is the “nearest”.
\emph{Meta-learning-based Approaches.}
Model-agnostic meta-learning~\cite{finn2017model} and LSTM-based meta-learner~\cite{ravi2016optimization} aim at searching for a set of good initialization weights, such that the classifier can rapidly generalize to novel tasks by fine-tuning on only a few support samples. 
\emph{Data-augmentation-based Approaches.}~\cite{hariharan2017low,wang2018low} generate synthetic examples of novel classes to improve the performance by using these synthetic examples for retraining. Our approach belongs to metric-learning-based approaches. It is the first framework towards CAPE. Besides, Keypoint Interaction Module (KIM) is specifically designed for CAPE to capture both the relationship among different keypoints and the relationship between support and query images.

\section{Class-Agnostic Pose Estimation (CAPE)}

\subsection{Problem Definition}

This paper introduces a novel task, termed class-agnostic pose estimation (CAPE). Unlike existing pose estimation tasks that predict keypoints of a single \emph{known/seen} (super-)category, CAPE requires a single model to detect keypoints of arbitrary category. More specifically, given one or a few support samples with keypoint definition of an \emph{unseen} category, object keypoints of this category can be detected without labeling large-scale supervisions and retraining models, significantly reducing the cost of data annotation and parameter tuning.

In order to validate the generalization capacity of CAPE models on unseen categories, they are trained on the \emph{base} categories but evaluated on \emph{novel} categories. The base categories and the novel categories are mutually exclusive, where the novel categories on the test set do not appear in the training data. During testing, CAPE models are provided with $K$ labeled support samples of an unseen category. The models are required to detect the poses of the query samples that are of the same category as the support samples. In this sense, CAPE task can be viewed as a K-shot pose estimation problem. Especially, when $K=1$, it is one-shot pose estimation.

\subsection{POse Matching Network (POMNet)}

\begin{figure}[t]
	\centering
	\includegraphics[width=0.98\textwidth]{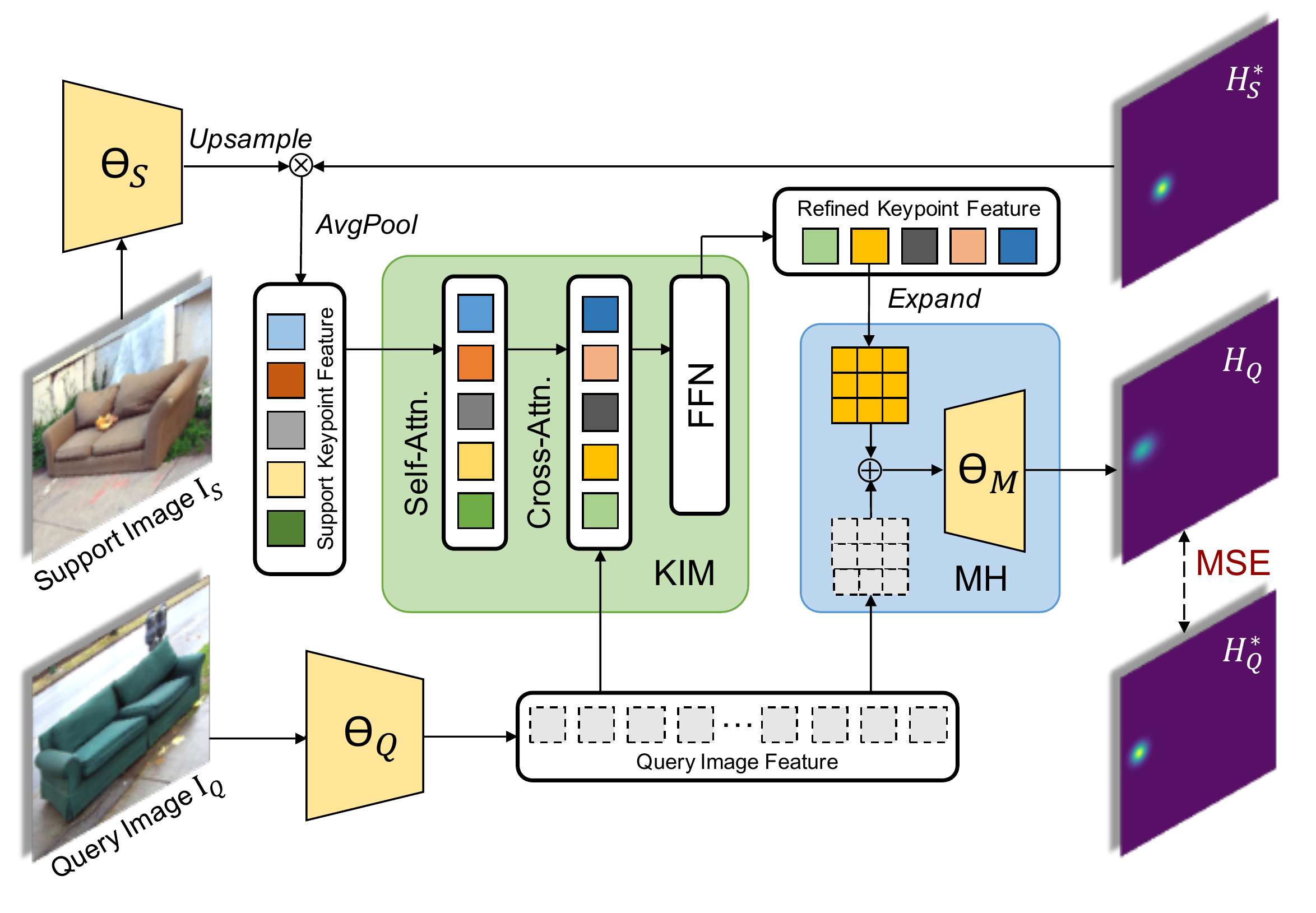}
	\caption{Overview of POse Matching Network (POMNet). Feature extractors $\mathrm{\Theta}_{S}$ and $\mathrm{\Theta}_{Q}$ extract the support keypoint features and the query image features respectively. Keypoint Interaction Module (KIM) refines the keypoint features by message passing among keypoints and capturing the relationship between the query and support images.
	Matching Head (MH) integrates the refined keypoint features and the query image features to predict the keypoint localization in the query image. MSE loss is applied to supervise the model.}
	\label{fig:transformer}
\end{figure}

Traditional pose estimators can be applied to neither the unseen object categories nor different keypoint definitions of the same class (\eg 19-keypoint human face definition and 68-keypoint human face definition). To achieve CAPE, we formulate the task as a matching problem and propose a novel framework termed POse Matching Network (POMNet). POMNet works by computing the matching similarity between the reference support keypoint features and the query image features at each location.
Therefore, POMNet is capable of handling various categories with different keypoint numbers and definitions.
As shown in Fig.~\ref{fig:transformer}, POMNet consists of three parts, \ie the feature extractors ($\mathrm{\Theta}_{S}$ and $\mathrm{\Theta}_{Q}$), Keypoint Interaction Module (KIM), and Matching Head (MH).

\textbf{Feature Extractor.}
We employ two parallel feature extractors to extract the support keypoint features and the query image features. In our implementation, ResNet-50~\cite{he2016deep} pre-trained on ImageNet dataset is used as the backbone.

For the support image $I_S$, the feature extractor $\mathrm{\Theta}_{S}$ is utilized to extract the support image features $\mathcal{F}_{S} = \mathrm{\Theta}_{S}(I_S).$ The keypoint annotations of the support sample are provided in the heatmap representations. We denote the ground-truth heatmaps of the support sample as ${H}_{S}^{*}$, and ${H}_{S}^{*j} \in \mathbb{R}^{H \times W \times 1}$ represents the heatmap of the $j_{th}$ keypoint. Given the support image features and the ground-truth heatmaps of the support sample, we can obtain the corresponding keypoint features as follows.

\begin{equation}
\begin{aligned}
    \hat{\mathcal{F}_{S}^{j}} = AvgPool(Upsample(\mathcal{F}_{S}) \otimes {H}_{S}^{*j}), \quad j=1, 2, ..., J
\end{aligned}
\label{eq:F_S}
\end{equation}
where ${\mathcal{F}_{S}} \in \mathbb{R}^{h \times w \times c}$ and $\hat{\mathcal{F}}_{S}^{j} \in \mathbb{R}^{1 \times 1 \times c}$ denote the support image features and the $j_{th}$ keypoint features respectively. $Upsample()$ is the up-sampling operation that reshapes the support image features to the same size of the corresponding heatmaps. $\otimes$ denotes pixel-wise multiplication. $AvgPool()$ is the average pooling operation that aggregates the support image features around the ground-truth keypoint position via weighted mean. $J$ is the number of reference keypoints.

For the query image $I_Q$, we follow a similar pipeline and apply the feature extractor $\mathrm{\Theta}_{Q}$ to extract the query image features $\mathcal{F}_{Q} = \mathrm{\Theta}_{Q}(I_Q)$. We collapse the spatial dimensions of the query image features and reshape them into a sequence. The extracted image features are then used to refine the support keypoint features in Keypoint Interaction Module (KIM) and to predict the keypoint localization in Matching Head (MH).

\textbf{Keypoint Interaction Module (KIM).}
KIM targets at enhancing the support keypoint features through efficient attention mechanisms. We first reduce the channel dimension of support keypoint features by a fully-connected layer and input the features of different keypoints as a sequence. As the keypoint numbers of different categories are different, several dummy features with padding mask are added at the end to keep a fixed number $L$ of input features ($L=100$ in our implementation), which enables KIM to adapt to various keypoint numbers. 
KIM has three transformer blocks, each of which consists of two major components, \ie Self-Attn. and Cross-Attn. \emph{Self-Attn.} The self-attention layer~\cite{vaswani2017attention} learns to exchange information among keypoints and utilize inherent object structures. It allows the keypoint features to interact with each other, and aggregate these interactions using the attention weights. \emph{Cross-Attn.} The keypoint features also interact with the query image features to align the feature representations and mitigate the representation gap. Specifically, a cross-attention layer~\cite{carion2020end} is applied to aggregate useful information in the query image. The keypoint features are input as query, and the flattened query image features are input as the key and the value. The channel dimension of the query image features are reduced to match the channel dimension of the keypoint features, and the sinusoidal position embedding~\cite{parmar2018image,vaswani2017attention} is supplemented to the query image features. A feed forward network (FFN) is also included following the common practice~\cite{vaswani2017attention}. As a result, the support keypoint features are processed and refined by KIM, $\{\bar{\mathcal{F}_{S}^{j}}\}_{j=1}^{L} = \text{KIM}(\{\hat{\mathcal{F}_{S}^{j}}\}_{j=1}^{L}, \mathcal{F}_{Q})$.We exclude the dummy padding ones and obtain the refined keypoint features $\{\bar{\mathcal{F}_{S}^{j}}\}_{j=1}^{J}$ by selecting the first $J$ valid keypoint features, where $J \leq L$.

\textbf{Matching Head (MH).} Given the refined keypoint features as the reference, Matching Head (MH) targets at seeking the best matching positions in the query image that are encoded with heatmaps.

We expand the refined keypoint features to the same spatial shape as the query image features $\mathcal{F}_{Q}$. The expanded features are then concatenated with the query image features. Finally, a decoder $\mathrm{\Theta}_{M}$ is employed to estimate the keypoint heatmaps. This procedure can be formulated as follows.

\begin{equation}
\begin{aligned}
    {H}_{Q}^{j} = \mathrm{\Theta}_{M}(Expand(\bar{\mathcal{F}_{S}^j}) \oplus \mathcal{F}_{Q}), \quad j = 1, 2, ..., J.
\end{aligned}
\label{eq:H_Q}
\end{equation}

\noindent where $\oplus$ refers to the channel-wise concatenation. $Expand()$ denotes the spatial expansion operation, \ie copying the refined keypoint features spatially to fit in the spatial size of the query image features. ${H}_{Q}^{j}$ is the predicted heatmap of the $j_{th}$ keypoint. The decoder  $\mathrm{\Theta}_{M}$ consists of one $3\times3$ convolutional layer, followed by deconvolutional layers for higher resolution as the common practice~\cite{xiao2018simple}. Pixel-wise mean squared error (MSE) loss is applied to supervise POMNet.

\begin{equation}
    \mathcal{L}_{MSE} = \dfrac{1}{JHW} \sum_{j=1}^J \sum_{\mathrm{p}} \lVert H_Q^j(\mathrm{p}) - H_Q^{*j}(\mathrm{p}) \rVert_2^2,
\label{eq:mse}
\end{equation}
\noindent where $H$ and $W$ refer to the height and width of heatmaps. $H_Q^j(\mathrm{p})$ and $H_Q^{*j}(\mathrm{p})$ are the predicted and the ground-truth pixel intensity at the position $\mathrm{p}$.
 
\textbf{Extension to K-shot.}
When $K$ ($K > 1$) support images are available, we first extract the support keypoint features for each sample individually, and then calculate the mean among the $K$ samples. The subsequent pipeline (including KIM and MH) is exactly the same as that of the 1-shot setting. With more support images, POMNet is able to capture more robust keypoint features to handle the intra-category variance and the ambiguity of the keypoint definition.

\section{Mulit-category Pose (MP-100) Dataset}

\begin{figure}[t]
	\centering
	\includegraphics[width=0.92\textwidth]{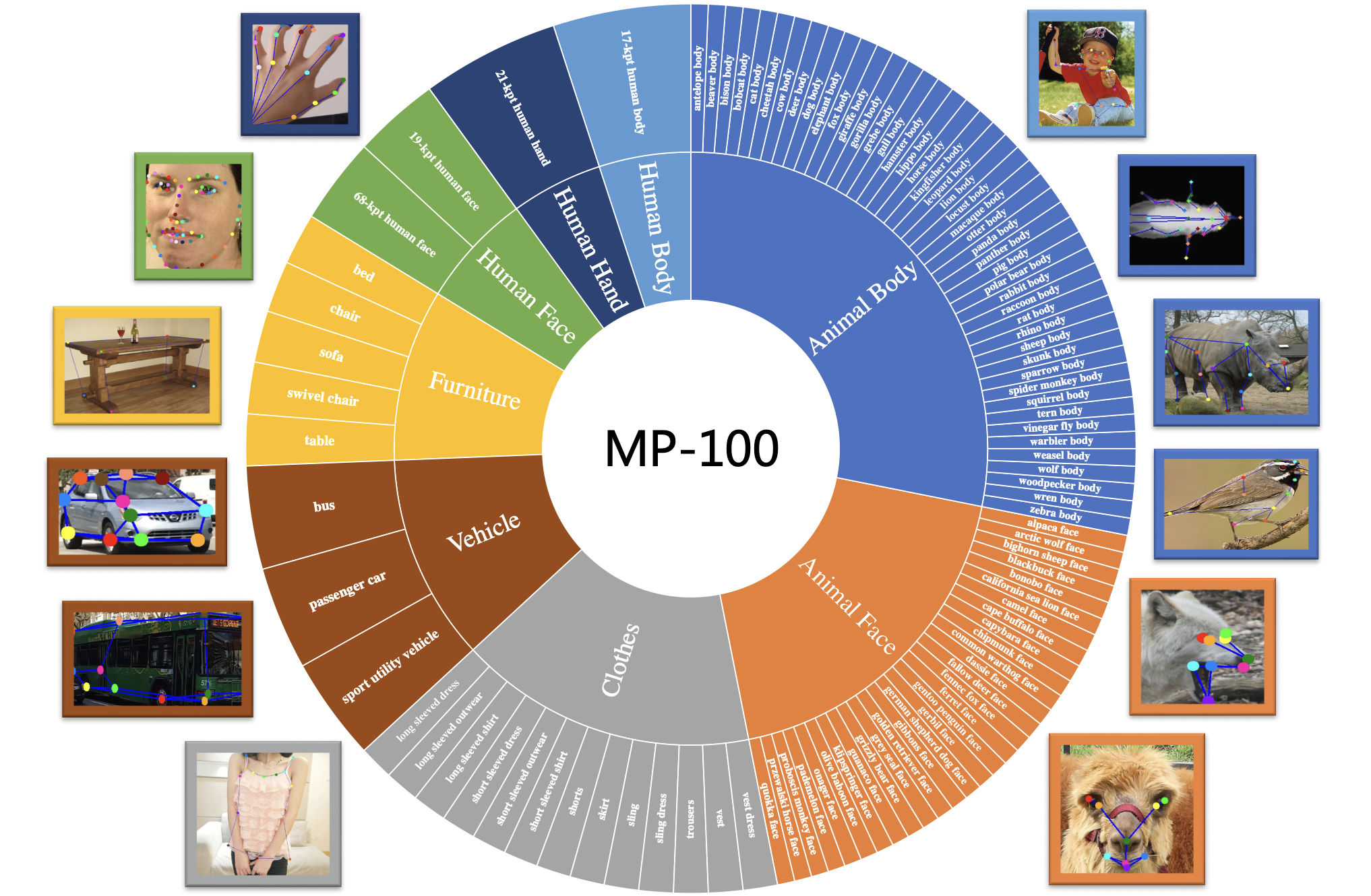}
	\caption{MP-100 dataset covers 100 sub-categories and 8 super-categories (human hand \& face \& body, animal face \& body, clothes, furniture, and vehicle).
	}
	\label{fig:MP}
\end{figure}

\begin{figure}[t]
	\centering
	\includegraphics[width=0.98\textwidth]{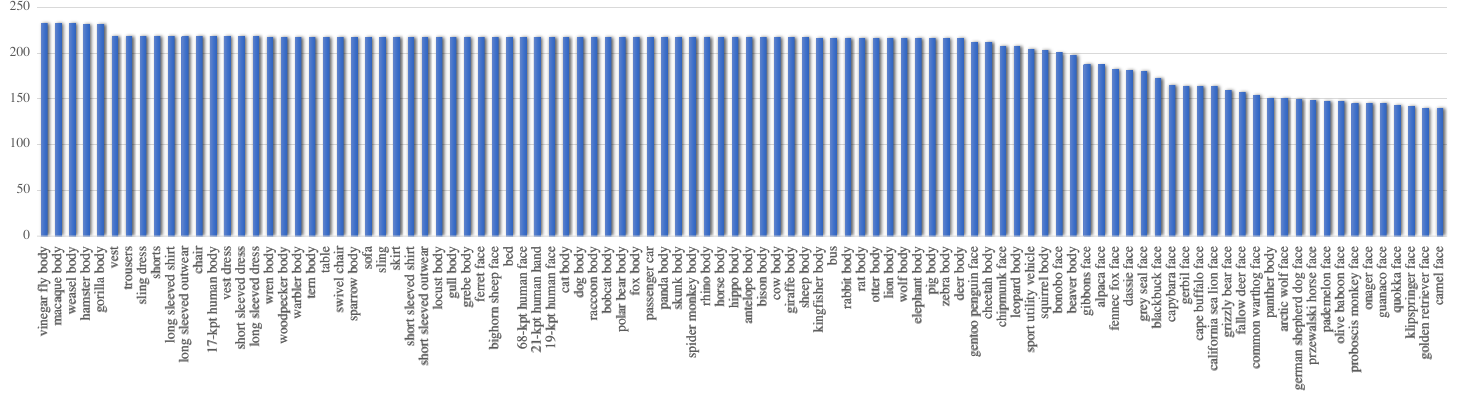}
	\caption{
	Histogram for instance number of each category on MP-100 dataset.
	}
	\label{fig:data_dist}
\end{figure}

Previous pose estimation datasets only consist of objects of one (super-)category and there are no existing datasets for CAPE task. We therefore construct the first large-scale pose dataset containing objects of multiple super-categories, termed Multi-category Pose (MP-100). In total, MP-100 dataset covers 100 sub-categories and 8 super-categories (human hand, human face, human body, animal body, animal face, clothes, furniture, and vehicle) as shown in Fig.~\ref{fig:MP}. 

Over 18K images and 20K annotations are collected from several popular 2D pose datasets, including COCO~\cite{lin2014microsoft}, 300W~\cite{300w}, AFLW~\cite{aflw}, OneHand10K~\cite{wang2018mask}, DeepFasion2~\cite{ge2019deepfashion2}, AP-10K~\cite{yu2021ap}, MacaquePose~\cite{labuguen2021macaquepose}, Vinegar Fly~\cite{pereira2019fast}, Desert Locust~\cite{graving2019deepposekit}, CUB-200~\cite{cub-200-2011}, CarFusion~\cite{reddy2018carfusion}, AnimalWeb~\cite{khan2020animalweb}, and Keypoint-5~\cite{wu2016single}. Keypoint numbers are diverse across different categories, ranging from 8 to 68.

We split the collected 100 categories into train/val/test sets (70 for train, 10 for val, and 20 for test). Following the common settings, we form five splits whose test sets are non-overlapping and evaluate the average model performance on the five splits. In this case, each category is tested as novel one on different splits and the category bias is avoided. Moreover, to treat all categories equally, we try our best to balance the number of instances among different categories. 

For the test set on each split, 2K samples are selected with 100 instances in each category. And for train/val set, 14K/2K samples are chosen for 70/10 categories respectively. As the number of instances available for different categories are extremely diverse and there are rare categories with less than 200 instances, we minimize the standard deviation of the instance number of all the categories. 
During sample selection, for each category, we give preference to the instances with more valid keypoints labeled and larger image resolution. 
In Fig.~\ref{fig:data_dist}, we demonstrate the histogram plot for the number of instances of each category on MP-100 dataset. The number of instances for each category is roughly balanced.

\section{Experiments}

\subsection{Implementation Details}

For each split on MP-100 dataset, we train POMNet on the train set, validate the performance on the val set, and finally evaluate the model on the test set. Note that the categories of the train/val/test set are mutually exclusive. During training, the support images and the query images of the same category are randomly paired. Each object of interest is cropped according to the bounding box and is resized to $256 \times 256$.
Data augmentation with random scaling ([$-15\%$, $15\%$]) and random rotation ([$-15^{\circ}$, $15^{\circ}$]) is applied to improve the model generalization ability. The training is conducted on 8 GPUs with a batch size of 16 in each GPU for 210 epochs. We follow MMPose~\cite{mmpose2020} to adopt Adam optimizer~\cite{kingma2014adam} with the base learning rate of 1e-3 and decay the learning rate to 1e-4 and 1e-5 respectively at the 170th and 200th epochs. During testing, we sample 3,000 random episodes for each novel category. Since there are 20 test categories for each split, we construct a total of 60,000 episodes for evaluation. 

PCK (Probability of Correct Keypoint) is a popular metric for pose estimation. If the normalized distance between the predicted keypoint and the ground-truth keypoint is less than a certain threshold ($\sigma$), it is considered correct. 

\begin{equation}
    \mathrm{PCK} = \frac{1}{N} \sum_{i=1}^{N} \mathbf{1} \left( \frac{ \lVert p_i - {p}_i^* \rVert_2}{d} \leq \sigma \right),
\end{equation}
where $p_i$ and ${p}_i^*$ are predicted and ground-truth keypoint locations respectively. $\mathbf{1}(\cdot)$ is the indicator function. $d$ is the longest side of the ground-truth bounding box, which is used as the normalization term. The correct ratio of the overall $N$ keypoints is calculated. 
In the experiments, we report the average PCK@0.2 ($\sigma$ = 0.2) of all the categories in each split. In order to minimize the category bias, the mean PCK result of all the 5 splits is also reported.

\subsection{Benchmark Results on MP-100 Dataset}

Class-Agnostic Pose Estimation (CAPE) is a new task that has not been tackled before. We tailor the existing few-shot learning baseline approaches, including Prototypical Networks~\cite{snell2017prototypical}, Finetune~\cite{nakamura2019revisiting}, and MAML~\cite{finn2017model} to address this new task. For fair comparisons, all baselines employ the same backbone network architecture (ResNet-50~\cite{he2016deep}) as ours.

\textbf{Prototypical Networks (ProtoNet)}~\cite{snell2017prototypical}. 
ProtoNet is a popular few-shot image classification approach, which constructs a prototype for each class and the query example is assigned to the class whose prototype is the ``nearest". To solve CAPE, we adapt ProtoNet to construct a prototype for each keypoint and find the location whose features are closest to the prototype in the query image. Unlike image classification, both the receptive fields and the spatial resolution of the features are critical for pose estimation. We empirically find that the features from Stage-3 achieve a good balance of these two factors among all the 4 stages.

\textbf{Fine-tune}~\cite{nakamura2019revisiting}. 
The model is first pre-trained using a combination of all base categories on the train set. During testing, the model is fine-tuned on the support images of the novel category before estimating the pose of the query images.
To handle the problem of various number of keypoints, the model is designed to output the maximum number of keypoints among all the categories, \ie 68 on MP-100 dataset, and only the few valid keypoints are supervised for each particular category during training and fine-tuning.

\textbf{Model-Agnostic Meta-Learning (MAML)}~\cite{finn2017model}.
Through meta training, the MAML model is explicitly trained to search for a good initialization such that its parameters can quickly adapt to the given category by fine-tuning on several support images. Similar to Fine-tune~\cite{nakamura2019revisiting}, the number of keypoints of the model is set as 68. During meta testing, the model can rapidly adapt to the novel categories given a few support images.

\begin{table*}[t]
    \setlength\tabcolsep{6pt}
	\caption{Comparisons with the baseline methods on MP-100 dataset under both 5-shot and 1-shot settings. POMNet significantly outperforms other approaches.}
	\label{tab:comparison}
	\begin{center}
	    \scalebox{0.95}{
		\begin{tabular}{c|ccccc|c}
		    \hline
			5-Shot & Split1 & Split2 & Split3 & Split4 & Split5 & Mean (PCK) \\ \hline
            ProtoNet~\cite{snell2017prototypical} & 60.31 & 53.51 & 61.92 & 58.44 & 58.61 & 58.56 \\
            MAML~\cite{finn2017model} & 70.03 & 55.98 & 63.21 & 64.79 & 58.47 & 62.50 \\
            Fine-tune~\cite{nakamura2019revisiting} & 71.67 & 57.84 & 66.76 & 66.53 & 60.24 & 64.61 \\
            POMNet (Ours) & \textbf{84.72} & \textbf{79.61} & \textbf{78.00} & \textbf{80.38} & \textbf{80.85} & \textbf{80.71} \\
			\hline
			\hline
			1-Shot & Split1 & Split2 & Split3 & Split4 & Split5 & Mean (PCK)\\ \hline
            ProtoNet~\cite{snell2017prototypical} & 46.05 & 40.84 & 49.13 & 43.34 & 44.54 & 44.78 \\
            MAML~\cite{finn2017model} & 68.14 & 54.72 & 64.19 & 63.24 & 57.20 & 61.50 \\
            Fine-tune~\cite{nakamura2019revisiting} & 70.60 & 57.04 & 66.06 & 65.00 & 59.20 & 63.58 \\
            POMNet (Ours) & \textbf{84.23} & \textbf{78.25} & \textbf{78.17} & \textbf{78.68} & \textbf{79.17} & \textbf{79.70} \\
			\hline
		\end{tabular}
		}
	\end{center}
\end{table*}

As shown in Table~\ref{tab:comparison}, our proposed POMNet shows superiority over the existing few-shot learning based approaches on the task of Class-Agnostic Pose Estimation (CAPE). We first conduct experimental comparisons under the 5-shot setting. 
We observe that ProtoNet~\cite{snell2017prototypical} mostly relies on low-level appearance features and encounters difficulties in constructing a reliable prototype using only 5 samples for all the keypoints. It processes each type of keypoint individually and does not utilize the structural information, restricting its upper bound performance.
MAML~\cite{finn2017model} and Fine-tune~\cite{nakamura2019revisiting} adapt to the novel object category by fine-tuning on a few samples during testing. However, the limited number of samples makes it hard for the model to achieve good performance on the novel categories due to severe over-fitting or under-fitting problems.
Our proposed POMNet considers the CAPE task as a matching problem, decoupling the model from the object category and the number of keypoints. In the meanwhile, KIM explicitly captures the relationship among keypoints and the structure of the object of interest. As a result, POMNet achieves 80.71 PCK on the novel categories under 5-shot settings, and outperforms the baseline methods by a large margin (over 25\% improvement). 

When the number of sample images decreases to one, the degeneration of our POMNet is only 1.3\% (79.70 vs 80.71 PCK). This is presumably because POMNet captures the relationship among semantic keypoints and is more robust to occlusion and visual ambiguity. In comparison, ProtoNet requires building the prototype based on a single keypoint, thus is more sensitive to the appearance variation, resulting in a larger performance drop (44.78 vs 58.46 PCK).

\subsection{Cross Super-Category Pose Estimation}

\begin{table*}[t]
    \setlength\tabcolsep{6pt}
	\caption{Cross super-category evaluation (PCK). POMNet outperforms other methods. But there is still large room for improvement on the rare categories.}
	\label{tab:cross-category}
	\begin{center}
	    \scalebox{0.95}{
		\begin{tabular}{c|cccc}
			\hline
			Method & Human Body & Human Face & Vehicle & Furniture \\ \hline
            ProtoNet~\cite{snell2017prototypical} & 37.61 & 57.80 & 28.35 & 42.64 \\
            MAML~\cite{finn2017model} & 51.93 & 25.72 & 17.68 & 20.09 \\
            Fine-tune~\cite{nakamura2019revisiting} & 52.11 & 25.53 & 17.46 & 20.76 \\
            POMNet (Ours) & \textbf{73.82} & \textbf{79.63} & \textbf{34.92} & \textbf{47.27} \\
			\hline
		\end{tabular}
		}
	\end{center}
\end{table*}

In order to further evaluate the generalization ability, we conduct the cross super-category pose estimation evaluation with the ``Leave-One-Out'' strategy.
That is, we train the model on all but one super-categories on MP-100 dataset, and evaluate the performance on the remaining one super-category. The super-categories to be evaluated include human body, human face, vehicle, and furniture.

As shown in Table~\ref{tab:cross-category}, our proposed POMNet outperforms the baseline methods on all the super-categories, demonstrating stronger generalization ability. However, super-category generalization is challenging and there is still a large room for improvement. We notice that all the methods perform poorly on the super-categories of vehicle and furniture. This is possibly because these categories are very different from the training ones and the extracted features are not discriminative enough. There are a great number of invisible keypoints for vehicle, and the intra-class variation between images is large for furniture, making these two super-categories more challenging. Solving CAPE requires to handle occlusion and intra-class appearance variation, and extract more discriminative features for unseen categories. We will explore these directions in the future.

\subsection{Ablation Study}

\begin{table*}[t]
    \setlength\tabcolsep{6pt}
	\caption{Ablation study of proposed components on MP-100 Split1 under 1-shot setting. KIM and MH significantly improve the model performance.}
	\label{tab:ablation}
	\begin{center}
	    \scalebox{0.95}{
		\begin{tabular}{l|ccc|c}
			\hline
			& Self-Atten. & Cross-Atten. & Matching Head & PCK \\ \hline
			\#1 &  &  & $\surd$ & 74.40 \\
		    \#2	& $\surd$ & $\surd$ &  & 79.19 \\
			\#3 &  & $\surd$ & $\surd$ & 80.76 \\
			\#4 & $\surd$ &  & $\surd$ & 82.92 \\
		    \#5 & $\surd$ & $\surd$ & $\surd$ & 84.23 \\
			\hline
		\end{tabular}
		}
	\end{center}
\end{table*}

\textbf{Effect of model components.} Table~\ref{tab:ablation} shows the effect of Keypoint Interaction Module (KIM) and Matching Head (MH). Comparing \#1 and \#5, we find that KIM significantly improves the CAPE model performance (13.2\% improvement). \#3 and \#4 show the effect of self-attention and cross-attention design, respectively. Especially, the 11.5\% gain from \#1 to \#4 shows that message passing among keypoints by self-attention greatly benefits keypoint localization. Comparison between \#2 and \#5 verifies the necessity of MH. \#2 replaces MH by matrix multiplication between support keypoint features and query image features. It collapses the channel dimension to 1 for each keypoint, causing undesirable information loss required for precise localization.

\begin{table}[htb]
    \setlength\tabcolsep{6pt}
	\begin{center}
	\caption{\textbf{Left:} Effect of training category number (``\#Train'') under 1-shot setting. Evaluation is conducted on a novel category (``human body").
	\textbf{Right:} Both training and testing are on ``human body" only.}
		\begin{tabular}{ccccc||cc}
			\hline
			\#Train & 1 & 9 & 49 & 99 & Oracle & SBL-Res50 [60] \\\hline
			PCK & 39.32 & 55.74 & 70.46 & 73.82 & 89.79 & 89.76 \\ \hline
		\end{tabular}
	\label{tab:num_class}
	\end{center}
\end{table}

\textbf{Effect of training category number.} As shown in Table~\ref{tab:num_class} \textbf{Left}, more training categories leads to better generalizability to the novel category, which validates the necessity of MP-100 dataset and the rationality of our experiments.

\textbf{Sanity check.} We perform traditional one class pose estimation as a sanity check. In Table~\ref{tab:num_class} \textbf{Right}, ``Oracle" means POMNet trained and tested on the same category (``human body") only. It performs comparable with SBL-Res50~\cite{he2016deep}, which demonstrates the correctness of our design choices.

\subsection{Qualitative Results}

\begin{figure}[t]
	\centering
	\includegraphics[width=0.88\textwidth]{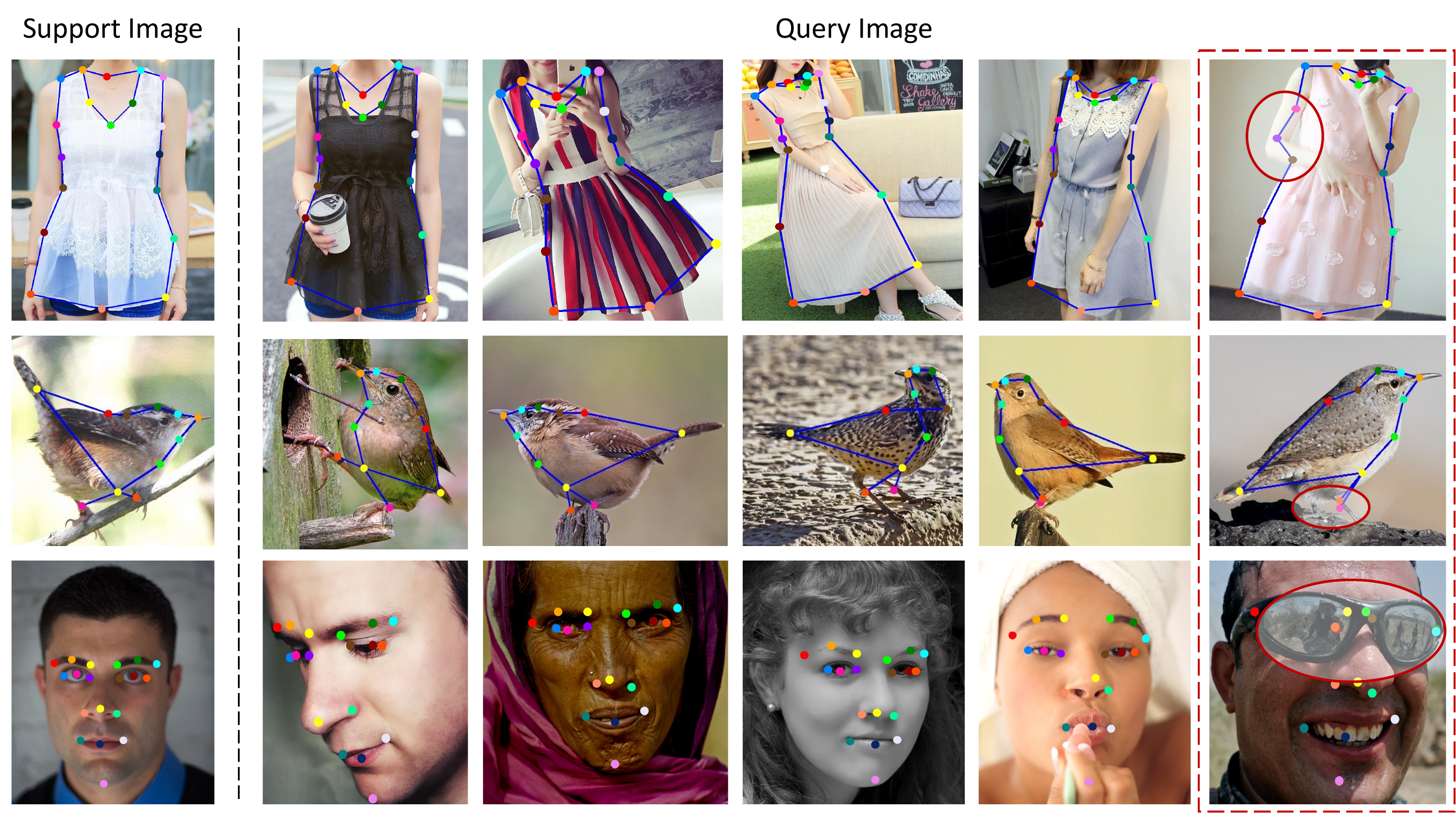}
	\caption{Qualitative results of POMNet on unseen categories. The first column shows the manually annotated support samples, and the others are the predicted query samples. The last column shows some failure cases (in \textcolor{red}{RED} circles).
	}
	\label{fig:viz}
\end{figure}

In Fig.~\ref{fig:viz}, we qualitatively evaluate the generalization ability of POMNet to the novel categories on MP-100 test sets. Our method is robust to perspective variation and appearance diversity. Typical failure cases include appearance ambiguity (the first two examples) and severe occlusion (the 3rd example).

\section{Conclusions and Limitations}

This paper introduces a novel task, termed Category-Agnostic Pose Estimation (CAPE). The idea of CAPE can benefit a wide range of application scenarios. It would not only promote the development of pose estimation (\eg pseudo-labeling for novel categories), but also enable the researchers in the other fields to detect keypoints of objects they are interested in (\eg plants). Besides, it may also make broader positive impacts for other computer vision tasks. For example, CAPE models can be developed for keypoint-based object tracking, contour-based instance segmentation, and graph matching. To achieve this goal, we propose the first CAPE framework, POse Matching Network (POMNet), and the first dataset for CAPE task, Multi-category Pose (MP-100). Experiments show that POMNet significantly outperforms the other approaches on MP-100 dataset. 
However, there are still many remaining challenges, \eg the generalization performance on rare categories, intra-class appearance variation, self-occlusion, and appearance ambiguity.
In conclusion, CAPE, as an important yet challenging task, is worth more research attention and further exploration.

~\\
\textbf{Acknowledgement.} 
This work is supported in part by the General Research Fund through the Research Grants Council of Hong Kong under Grants (Nos. 14202217, 14203118, 14208619), in part by Research Impact Fund Grant No. R5001-18.
Ping Luo is supported by the General Research Fund of HK No.27208720, No.17212120, and No.17200622.
Wanli Ouyang is supported by the Australian Research Council Grant DP200103223, Australian Medical Research Future Fund MRFAI000085, CRC-P Smart Material Recovery Facility (SMRF) – Curby Soft Plastics, and CRC-P ARIA - Bionic Visual-Spatial Prosthesis for the Blind.

\clearpage
% ---- Bibliography ----
%
% BibTeX users should specify bibliography style 'splncs04'.
% References will then be sorted and formatted in the correct style.
%
\bibliographystyle{splncs04}
\bibliography{egbib}

\begin{thebibliography}{10}
\providecommand{\url}[1]{\texttt{#1}}
\providecommand{\urlprefix}{URL }
\providecommand{\doi}[1]{https://doi.org/#1}

\bibitem{andriluka2018posetrack}
Andriluka, M., Iqbal, U., Insafutdinov, E., Pishchulin, L., Milan, A., Gall,
  J., Schiele, B.: Posetrack: A benchmark for human pose estimation and
  tracking. In: IEEE Conf. Comput. Vis. Pattern Recog. (2018)

\bibitem{mpii}
Andriluka, M., Pishchulin, L., Gehler, P., Schiele, B.: 2d human pose
  estimation: New benchmark and state of the art analysis. In: IEEE Conf.
  Comput. Vis. Pattern Recog. (2014)

\bibitem{bulat2017far}
Bulat, A., Tzimiropoulos, G.: How far are we from solving the 2d \& 3d face
  alignment problem? In: Int. Conf. Comput. Vis. (2017)

\bibitem{cao2019cross}
Cao, J., Tang, H., Fang, H.S., Shen, X., Lu, C., Tai, Y.W.: Cross-domain
  adaptation for animal pose estimation. In: Int. Conf. Comput. Vis. (2019)

\bibitem{carion2020end}
Carion, N., Massa, F., Synnaeve, G., Usunier, N., Kirillov, A., Zagoruyko, S.:
  End-to-end object detection with transformers. In: Eur. Conf. Comput. Vis.
  (2020)

\bibitem{chen2018cascaded}
Chen, Y., Wang, Z., Peng, Y., Zhang, Z., Yu, G., Sun, J.: Cascaded pyramid
  network for multi-person pose estimation. In: IEEE Conf. Comput. Vis. Pattern
  Recog. (2018)

\bibitem{cheng2020higherhrnet}
Cheng, B., Xiao, B., Wang, J., Shi, H., Huang, T.S., Zhang, L.: Higherhrnet:
  Scale-aware representation learning for bottom-up human pose estimation. In:
  IEEE Conf. Comput. Vis. Pattern Recog. (2020)

\bibitem{chu2017multi}
Chu, X., Yang, W., Ouyang, W., Ma, C., Yuille, A.L., Wang, X.: Multi-context
  attention for human pose estimation. In: IEEE Conf. Comput. Vis. Pattern
  Recog. (2017)

\bibitem{mmpose2020}
Contributors, M.: Openmmlab pose estimation toolbox and benchmark.
  \url{https://github.com/open-mmlab/mmpose} (2020)

\bibitem{duan2019trb}
Duan, H., Lin, K.Y., Jin, S., Liu, W., Qian, C., Ouyang, W.: Trb: a novel
  triplet representation for understanding 2d human body. In: Int. Conf.
  Comput. Vis. (2019)

\bibitem{finn2017model}
Finn, C., Abbeel, P., Levine, S.: Model-agnostic meta-learning for fast
  adaptation of deep networks. In: ICML (2017)

\bibitem{ge2019deepfashion2}
Ge, Y., Zhang, R., Wang, X., Tang, X., Luo, P.: Deepfashion2: A versatile
  benchmark for detection, pose estimation, segmentation and re-identification
  of clothing images. In: IEEE Conf. Comput. Vis. Pattern Recog. (2019)

\bibitem{graving2019deepposekit}
Graving, J.M., Chae, D., Naik, H., Li, L., Koger, B., Costelloe, B.R., Couzin,
  I.D.: Deepposekit, a software toolkit for fast and robust animal pose
  estimation using deep learning. Elife  (2019)

\bibitem{hariharan2017low}
Hariharan, B., Girshick, R.: Low-shot visual recognition by shrinking and
  hallucinating features. In: Int. Conf. Comput. Vis. (2017)

\bibitem{he2016deep}
He, K., Zhang, X., Ren, S., Sun, J.: Deep residual learning for image
  recognition. In: IEEE Conf. Comput. Vis. Pattern Recog. (2016)

\bibitem{Jhuang2013}
Jhuang, H., Gall, J., Zuffi, S., Schmid, C., Black, M.J.: Towards understanding
  action recognition. In: Int. Conf. Comput. Vis. (2013)

\bibitem{jiang2019class}
Jiang, S., Liang, S., Chen, C., Zhu, Y., Li, X.: Class agnostic image common
  object detection. IEEE Trans. Image Process.  \textbf{28}(6),  2836--2846
  (2019)

\bibitem{jin2019multi}
Jin, S., Liu, W., Ouyang, W., Qian, C.: Multi-person articulated tracking with
  spatial and temporal embeddings. In: IEEE Conf. Comput. Vis. Pattern Recog.
  (2019)

\bibitem{jin2020differentiable}
Jin, S., Liu, W., Xie, E., Wang, W., Qian, C., Ouyang, W., Luo, P.:
  Differentiable hierarchical graph grouping for multi-person pose estimation.
  In: Eur. Conf. Comput. Vis. (2020)

\bibitem{jin2017towards}
Jin, S., Ma, X., Han, Z., Wu, Y., Yang, W., Liu, W., Qian, C., Ouyang, W.:
  Towards multi-person pose tracking: Bottom-up and top-down methods. In: Int.
  Conf. Comput. Vis. Worksh. (2017)

\bibitem{jin2020whole}
Jin, S., Xu, L., Xu, J., Wang, C., Liu, W., Qian, C., Ouyang, W., Luo, P.:
  Whole-body human pose estimation in the wild. In: Eur. Conf. Comput. Vis.
  (2020)

\bibitem{khan2020animalweb}
Khan, M.H., McDonagh, J., Khan, S., Shahabuddin, M., Arora, A., Khan, F.S.,
  Shao, L., Tzimiropoulos, G.: Animalweb: A large-scale hierarchical dataset of
  annotated animal faces. In: IEEE Conf. Comput. Vis. Pattern Recog. (2020)

\bibitem{kingma2014adam}
Kingma, D.P., Ba, J.: Adam: A method for stochastic optimization. In: Int.
  Conf. Learn. Represent. (2015)

\bibitem{aflw}
Kostinger, M., Wohlhart, P., Roth, P., Bischof, H.: Annotated facial landmarks
  in the wild: A large-scale, real-world database for facial landmark
  localization. In: Int. Conf. Comput. Vis. Worksh. (2011)

\bibitem{labuguen2021macaquepose}
Labuguen, R., Matsumoto, J., Negrete, S.B., Nishimaru, H., Nishijo, H., Takada,
  M., Go, Y., Inoue, K.i., Shibata, T.: Macaquepose: A novel “in the wild”
  macaque monkey pose dataset for markerless motion capture. Frontiers in
  behavioral neuroscience  (2021)

\bibitem{li2021human}
Li, J., Bian, S., Zeng, A., Wang, C., Pang, B., Liu, W., Lu, C.: Human pose
  regression with residual log-likelihood estimation. In: Int. Conf. Comput.
  Vis. (2021)

\bibitem{li2019crowdpose}
Li, J., Wang, C., Zhu, H., Mao, Y., Fang, H.S., Lu, C.: Crowdpose: Efficient
  crowded scenes pose estimation and a new benchmark. In: IEEE Conf. Comput.
  Vis. Pattern Recog. (2019)

\bibitem{li2020atrw}
Li, S., Li, J., Tang, H., Qian, R., Lin, W.: Atrw: A benchmark for amur tiger
  re-identification in the wild. In: ACM Int. Conf. Multimedia (2020)

\bibitem{li2021tokenpose}
Li, Y., Zhang, S., Wang, Z., Yang, S., Yang, W., Xia, S.T., Zhou, E.:
  Tokenpose: Learning keypoint tokens for human pose estimation. arXiv preprint
  arXiv:2104.03516  (2021)

\bibitem{lin2014microsoft}
Lin, T.Y., Maire, M., Belongie, S., Hays, J., Perona, P., Ramanan, D.,
  Doll{\'a}r, P., Zitnick, C.L.: Microsoft coco: Common objects in context. In:
  Eur. Conf. Comput. Vis. (2014)

\bibitem{lu2018class}
Lu, E., Xie, W., Zisserman, A.: Class-agnostic counting. In: ACCV (2018)

\bibitem{lu2020learning}
Lu, J., Gong, P., Ye, J., Zhang, C.: Learning from very few samples: A survey.
  arXiv preprint arXiv:2009.02653  (2020)

\bibitem{mao2021tfpose}
Mao, W., Ge, Y., Shen, C., Tian, Z., Wang, X., Wang, Z.: Tfpose: Direct human
  pose estimation with transformers. arXiv preprint arXiv:2103.15320  (2021)

\bibitem{mathis2021pretraining}
Mathis, A., Biasi, T., Schneider, S., Yuksekgonul, M., Rogers, B., Bethge, M.,
  Mathis, M.W.: Pretraining boosts out-of-domain robustness for pose
  estimation. In: Proceedings of the IEEE/CVF Winter Conference on Applications
  of Computer Vision (2021)

\bibitem{Moon_2020_ECCV_InterHand2.6M}
Moon, G., Yu, S.I., Wen, H., Shiratori, T., Lee, K.M.: Interhand2.6m: A dataset
  and baseline for 3d interacting hand pose estimation from a single rgb image.
  In: Eur. Conf. Comput. Vis. (2020)

\bibitem{mueller2018ganerated}
Mueller, F., Bernard, F., Sotnychenko, O., Mehta, D., Sridhar, S., Casas, D.,
  Theobalt, C.: Ganerated hands for real-time 3d hand tracking from monocular
  rgb. In: IEEE Conf. Comput. Vis. Pattern Recog. (2018)

\bibitem{mueller2017real}
Mueller, F., Mehta, D., Sotnychenko, O., Sridhar, S., Casas, D., Theobalt, C.:
  Real-time hand tracking under occlusion from an egocentric rgb-d sensor. In:
  Int. Conf. Comput. Vis. (2017)

\bibitem{nakamura2019revisiting}
Nakamura, A., Harada, T.: Revisiting fine-tuning for few-shot learning. arXiv
  preprint arXiv:1910.00216  (2019)

\bibitem{newell2016stacked}
Newell, A., Yang, K., Deng, J.: Stacked hourglass networks for human pose
  estimation. In: Eur. Conf. Comput. Vis. (2016)

\bibitem{nie2019single}
Nie, X., Feng, J., Zhang, J., Yan, S.: Single-stage multi-person pose machines.
  In: Int. Conf. Comput. Vis. (2019)

\bibitem{parmar2018image}
Parmar, N., Vaswani, A., Uszkoreit, J., Kaiser, L., Shazeer, N., Ku, A., Tran,
  D.: Image transformer. In: ICML (2018)

\bibitem{pereira2019fast}
Pereira, T.D., Aldarondo, D.E., Willmore, L., Kislin, M., Wang, S.S.H., Murthy,
  M., Shaevitz, J.W.: Fast animal pose estimation using deep neural networks.
  Nature methods  (2019)

\bibitem{ravi2016optimization}
Ravi, S., Larochelle, H.: Optimization as a model for few-shot learning. In:
  Int. Conf. Learn. Represent. (2017)

\bibitem{reddy2018carfusion}
Reddy, N.D., Vo, M., Narasimhan, S.G.: Carfusion: Combining point tracking and
  part detection for dynamic 3d reconstruction of vehicles. In: IEEE Conf.
  Comput. Vis. Pattern Recog. (2018)

\bibitem{300w}
Sagonas, C., Antonakos, E., Tzimiropoulos, G., Zafeiriou, S., Pantic, M.: 300
  faces in-the-wild challenge: database and results. Image and Vision Computing
   (2016)

\bibitem{shen2015first}
Shen, J., Zafeiriou, S., Chrysos, G.G., Kossaifi, J., Tzimiropoulos, G.,
  Pantic, M.: The first facial landmark tracking in-the-wild challenge:
  Benchmark and results. In: Int. Conf. Comput. Vis. Worksh. (2015)

\bibitem{simon2017hand}
Simon, T., Joo, H., Matthews, I., Sheikh, Y.: Hand keypoint detection in single
  images using multiview bootstrapping. In: IEEE Conf. Comput. Vis. Pattern
  Recog. (2017)

\bibitem{snell2017prototypical}
Snell, J., Swersky, K., Zemel, R.: Prototypical networks for few-shot learning.
  Adv. Neural Inform. Process. Syst.  (2017)

\bibitem{song2019apollocar3d}
Song, X., Wang, P., Zhou, D., Zhu, R., Guan, C., Dai, Y., Su, H., Li, H., Yang,
  R.: Apollocar3d: A large 3d car instance understanding benchmark for
  autonomous driving. In: IEEE Conf. Comput. Vis. Pattern Recog. (2019)

\bibitem{sun2019deep}
Sun, K., Xiao, B., Liu, D., Wang, J.: Deep high-resolution representation
  learning for human pose estimation. In: IEEE Conf. Comput. Vis. Pattern
  Recog. (2019)

\bibitem{sun2017compositional}
Sun, X., Shang, J., Liang, S., Wei, Y.: Compositional human pose regression.
  In: Int. Conf. Comput. Vis. (2017)

\bibitem{toshev2014deeppose}
Toshev, A., Szegedy, C.: Deeppose: Human pose estimation via deep neural
  networks. In: IEEE Conf. Comput. Vis. Pattern Recog. (2014)

\bibitem{vaswani2017attention}
Vaswani, A., Shazeer, N., Parmar, N., Uszkoreit, J., Jones, L., Gomez, A.N.,
  Kaiser, {\L}., Polosukhin, I.: Attention is all you need. Adv. Neural Inform.
  Process. Syst.  (2017)

\bibitem{vinyals2016matching}
Vinyals, O., Blundell, C., Lillicrap, T., Wierstra, D., et~al.: Matching
  networks for one shot learning. In: Adv. Neural Inform. Process. Syst. (2016)

\bibitem{wang2018mask}
Wang, Y., Peng, C., Liu, Y.: Mask-pose cascaded cnn for 2d hand pose estimation
  from single color image. IEEE Transactions on Circuits and Systems for Video
  Technology  (2018)

\bibitem{wang2018low}
Wang, Y.X., Girshick, R., Hebert, M., Hariharan, B.: Low-shot learning from
  imaginary data. In: IEEE Conf. Comput. Vis. Pattern Recog. (2018)

\bibitem{wei2016convolutional}
Wei, S.E., Ramakrishna, V., Kanade, T., Sheikh, Y.: Convolutional pose
  machines. In: IEEE Conf. Comput. Vis. Pattern Recog. (2016)

\bibitem{cub-200-2011}
Welinder, P., Branson, S., Mita, T., Wah, C., Schroff, F., Belongie, S.,
  Perona, P.: {Caltech-UCSD Birds 200}. Tech. Rep. CNS-TR-2010-001, California
  Institute of Technology (2010)

\bibitem{wu2017ai}
Wu, J., Zheng, H., Zhao, B., Li, Y., Yan, B., Liang, R., Wang, W., Zhou, S.,
  Lin, G., Fu, Y., et~al.: Ai challenger: A large-scale dataset for going
  deeper in image understanding. arXiv preprint arXiv:1711.06475  (2017)

\bibitem{wu2016single}
Wu, J., Xue, T., Lim, J.J., Tian, Y., Tenenbaum, J.B., Torralba, A., Freeman,
  W.T.: Single image 3d interpreter network. In: Eur. Conf. Comput. Vis. (2016)

\bibitem{wu2018look}
Wu, W., Qian, C., Yang, S., Wang, Q., Cai, Y., Zhou, Q.: Look at boundary: A
  boundary-aware face alignment algorithm. In: IEEE Conf. Comput. Vis. Pattern
  Recog. (2018)

\bibitem{xiao2018simple}
Xiao, B., Wu, H., Wei, Y.: Simple baselines for human pose estimation and
  tracking. In: Eur. Conf. Comput. Vis. (2018)

\bibitem{xu2021vipnas}
Xu, L., Guan, Y., Jin, S., Liu, W., Qian, C., Luo, P., Ouyang, W., Wang, X.:
  Vipnas: Efficient video pose estimation via neural architecture search. In:
  IEEE Conf. Comput. Vis. Pattern Recog. (2021)

\bibitem{yang2018learning}
Yang, F.S.Y., Zhang, L., Xiang, T., Torr, P.H., Hospedales, T.M.: Learning to
  compare: Relation network for few-shot learning. In: IEEE Conf. Comput. Vis.
  Pattern Recog. (2018)

\bibitem{yang2020transpose}
Yang, S., Quan, Z., Nie, M., Yang, W.: Transpose: Towards explainable human
  pose estimation by transformer. arXiv preprint arXiv:2012.14214  (2020)

\bibitem{yang2021class}
Yang, S.D., Su, H.T., Hsu, W.H., Chen, W.C.: Class-agnostic few-shot object
  counting. In: Proceedings of the IEEE/CVF Winter Conference on Applications
  of Computer Vision (2021)

\bibitem{yu2021ap}
Yu, H., Xu, Y., Zhang, J., Zhao, W., Guan, Z., Tao, D.: Ap-10k: A benchmark for
  animal pose estimation in the wild. arXiv preprint arXiv:2108.12617  (2021)

\bibitem{yuan2021hrformer}
Yuan, Y., Fu, R., Huang, L., Lin, W., Zhang, C., Chen, X., Wang, J.: Hrformer:
  High-resolution transformer for dense prediction. arXiv preprint
  arXiv:2110.09408  (2021)

\bibitem{zafeiriou2017menpo}
Zafeiriou, S., Trigeorgis, G., Chrysos, G., Deng, J., Shen, J.: The menpo
  facial landmark localisation challenge: A step towards the solution. In: IEEE
  Conf. Comput. Vis. Pattern Recog. Worksh. (2017)

\bibitem{zeng2022not}
Zeng, W., Jin, S., Liu, W., Qian, C., Luo, P., Ouyang, W., Wang, X.: Not all
  tokens are equal: Human-centric visual analysis via token clustering
  transformer. In: IEEE Conf. Comput. Vis. Pattern Recog. (2022)

\bibitem{zhang2019canet}
Zhang, C., Lin, G., Liu, F., Yao, R., Shen, C.: Canet: Class-agnostic
  segmentation networks with iterative refinement and attentive few-shot
  learning. In: IEEE Conf. Comput. Vis. Pattern Recog. pp. 5217--5226 (2019)

\bibitem{zhang2019pose2seg}
Zhang, S.H., Li, R., Dong, X., Rosin, P., Cai, Z., Han, X., Yang, D., Huang,
  H., Hu, S.M.: Pose2seg: Detection free human instance segmentation. In: IEEE
  Conf. Comput. Vis. Pattern Recog. (2019)

\bibitem{zhao2018understanding}
Zhao, J., Li, J., Cheng, Y., Sim, T., Yan, S., Feng, J.: Understanding humans
  in crowded scenes: Deep nested adversarial learning and a new benchmark for
  multi-human parsing. In: ACM Int. Conf. Multimedia (2018)

\bibitem{zhou2018starmap}
Zhou, X., Karpur, A., Luo, L., Huang, Q.: Starmap for category-agnostic
  keypoint and viewpoint estimation. In: Eur. Conf. Comput. Vis. pp. 318--334
  (2018)

\bibitem{zimmermann2017learning}
Zimmermann, C., Brox, T.: Learning to estimate 3d hand pose from single rgb
  images. In: Int. Conf. Comput. Vis. (2017)

\bibitem{zimmermann2019freihand}
Zimmermann, C., Ceylan, D., Yang, J., Russell, B., Argus, M., Brox, T.:
  Freihand: A dataset for markerless capture of hand pose and shape from single
  rgb images. In: Int. Conf. Comput. Vis. (2019)

\end{thebibliography}
\end{document}